\documentclass[preprint,12pt]{elsarticle}

\usepackage{amssymb}
\usepackage{amsmath}
\usepackage{graphicx}
\usepackage{epsfig}
\usepackage{subcaption}
\usepackage{hyperref}
\journal{arXiv}

\begin{document}

\begin{frontmatter}




\title{Controllable Edge-Type-Specific Interpretation in Multi-Relational Graph Neural Networks for Drug Response Prediction}

\author[university]{Xiaodi Li\fnref{equal}}
\ead{lixiaodi@stu.ahau.edu.cn}

\author[university]{Jianfeng Gui\fnref{equal}}
\ead{gjf2fm@stu.ahau.edu.cn}

\author[university]{Qian Gao}
\ead{gaoqian@stu.ahau.edu.cn}

\author[university]{Haoyuan Shi}
\ead{haoyuanshi@stu.ahau.edu.cn}

\author[university]{Zhenyu Yue\corref{corresponding}}
\ead{zhenyuyue@ahau.edu.cn}
\ead[url]{https://orcid.org/0000-0002-9370-2540}

\cortext[corresponding]{Corresponding author.}
\fntext[equal]{The two authors contribute equally to this work.}
\affiliation[university]{organization={School of Information and Artificial Intelligence, Anhui Agricultural University},
            addressline={Changjiang West Road}, 
            city={Hefei},
            postcode={230036}, 
            state={East Asia},
            country={China}}


\begin{abstract}
Graph Neural Networks have been widely applied in critical decision-making areas that demand interpretable predictions, leading to the flourishing development of interpretability algorithms. However, current graph interpretability algorithms tend to emphasize generality and often overlook biological significance, thereby limiting their applicability in predicting cancer drug responses. In this paper, we propose a novel post-hoc interpretability algorithm for cancer drug response prediction, CETExplainer, which incorporates a controllable edge-type-specific weighting mechanism. It considers the mutual information between subgraphs and predictions, proposing a structural scoring approach to provide fine-grained, biologically meaningful explanations for predictive models. We also introduce a method for constructing ground truth based on real-world datasets to quantitatively evaluate the proposed interpretability algorithm. Empirical analysis on the real-world dataset demonstrates that CETExplainer achieves superior stability and improves explanation quality compared to leading algorithms, thereby offering a robust and insightful tool for cancer drug prediction. 
\end{abstract}


\end{frontmatter}


\section{Introduction}
\label{sec1}
Precisely identifying cancer drug response (CDR) holds great promise for developing personalized therapy, which can increase survival rates and reduce expenses \cite{adamMachineLearningApproaches2020,shen2021dbmcs}. Since testing multiple drugs for a cancer patient is infeasible for practical and financial reasons, there is an urgent demand for computational methods that can accurately predict CDR \cite{leighl2021arm,xuReusabilityReportUncovering2024}.

Recently, the developments of Graph Neural Networks (GNNs) have revolutionized the bioinformatics fields, especially in CDR prediction tasks \cite{shen2022pharmacogenomic}. For example, MOFGCN leverages GNNs to extract valuable information from a heterogeneous network comprising both cell line and drug nodes to predict drug sensitivity \cite{pengPredictingDrugResponse2022}. DualGCN combines the chemical structures of drugs and omics data of biological samples to predict CDR through a Dual Graph Convolutional Network model and achieved promising results \cite{maDualGCNDualGraph2022}. However, due to the lack of interpretability considerations during the construction of GNN models, they are often regarded as ‘black boxes’, which undermine the trust of physicians and patients in their predictions \cite{amannExplainabilityArtificialIntelligence2020}. Fortunately, numerous interpretability algorithms for GNNs have been developed \cite{kakkadSurveyExplainabilityGraph2023}. For example, ExplaiNE \cite{kangExplaiNEApproachExplaining2019a} employs counterfactual explanations, which involve interpreting predictions by quantifying how weakening an existing link affects the prediction probability. GNNExplainer \cite{ying2019gnnexplainer} and PGExplainer \cite{luoInductiveEfficientExplanations2024} both treat the soft mask as a trainable variate, selecting a concise subgraph and maximizing the mutual information between the subgraph and the prediction as the explanation. However, when applying these explainability methods to CDR data, we encountered three challenges. 1. When CDR data are integrated with cell line multi-omics data and drug molecular fingerprint data, they create high-dimensional and highly complex datasets \cite{baptistaDeepLearningDrug2021}. Effectively extracting feature information among nodes within these datasets presents a significant challenge. 2. Current perturbation-based explainability algorithm only focuses on the smallest subgraph that most impacts the prediction, ignoring the fact that the explanation subgraph should contain specific types of edges, which are important in the bioinformatics field. 3. In real-world datasets, the explanations generated by interpretability algorithms often lack definitive ground truth (GT). Developing a universally accepted method to construct GT using specialized biological knowledge remains a significant challenge.

To tackle these challenges, we propose a controllable edge-type-specific weight mechanism for interpretation in multi-relational graph neural networks (CETExplainer) to explore the topological relationships between cancer cell lines and drugs. Specifically, according to \cite{deng2021xgraphboost}, directed graph neural networks demonstrate superior capabilities in molecular feature extraction compared to undirected graph neural networks. To address challenge 1, we treat cell lines and drugs as nodes, with the interactions between them as directed edges, constructing a directed heterogeneous network. We then utilize relational graph convolutional network (R-GCN) layers \cite{schlichtkrull2018modeling} to extract molecule-level features from the cell line and drug nodes for CDR prediction. For challenge 2, we develop an algorithm that leverages mutual information maximization between subgraphs and predictions, and incorporates subgraph structural scoring to enhance biological significance and improve explanation quality. To address the lack of reference benchmarks in explaining real world data, we establish a method to construct GT and then propose three metrics to use GT to assess the effectiveness of model explanations and solve the challenge 3. There are the main contributions of this work. (i) We treat the CDR data, which includes sensitive and resistant edges, from the perspective of a directed heterogeneous graph. Using a GNN model, we encode node information to perform link prediction, thereby uncovering potential drug response relationships. (ii) We propose a controllable edge-type-specific interpretability mechanism that can assign attention to various edges, thereby enabling the model to focus more on meaningful and easily interpretable structures. (iii) We create GT for CETExplainer and propose three metrics to quantitatively evaluate the performance of explainable algorithms.

\section{Related work}
\label{sec2}

\subsection{Interpretation method for drug responses}
\label{subsec2.1}
Previous researches have made significant attempts in the domain of CDR prediction and its interpretation. TANDEM \cite{abenTANDEMTwostageApproach2016} employs Elastic Net regression, a linear model known for its interpretability. Each coefficient in the model reflects the importance of the corresponding feature in the prediction, enabling researchers to intuitively understand which biomarkers are significantly associated with CDR. However, linear models typically struggle to capture the complex interactions between variables, which may limit their ability to characterize the complexity of biological systems effectively. PathDSP \cite{tang2021explainable} and BDKANN \cite{snowInterpretableDrugResponse2021} utilized deep learning techniques to predict CDR and use SHAP values to quantify the contribution of each feature to the model's predictions, thereby enhancing transparency and offering deeper insights into the decision-making process. SHAP values can provide explanations for predictions from separable feature data. However, for models that integrate multi-dimensional data, the fused features are often highly correlated and lack separability, which poses challenges to the effective application of SHAP values \cite{aas2021explaining}. SubCDR \cite{liuSubcomponentguidedDeepLearning2023} leveraged graph convolution networks, which inherently excel at processing structured data, particularly in analyzing interactions and network structures, to address the challenge of predicting CDR. SubCDR demonstrated the influence of key subcomponents through interaction graphs, thereby offering an intuitive interpretation of the model’s decisions. However, the internal mechanisms of GNNs remained complex and may require specialized knowledge for full comprehension. 
Subsection text.
\subsection{Interpretation algorithm for GNN}
\label{subsec2.2}
Although existing interpretative approaches for CDR prediction have achieved commendable results, the methods used may not meet the evolving interpretative needs in the CDR field, especially as an increasing number of researchers employ GNNs to address problems in this area. Fortunately, several interpretability algorithms have been proposed, offering diverse solutions. For example, GNNExplainer \cite{ying2019gnnexplainer} optimized a mask to identify minimal explanatory subgraphs based on perturbations; the proxy-based approach GraphLIME \cite{huangGraphLIMELocalInterpretable2020} utilized the HSIC Lasso non-linear proxy model for interpretations; Excitation-BP \cite{popeExplainabilityMethodsGraph2019} decomposed the target probability into several conditional probability terms for explanations; gradient-based saliency methods SA \cite{baldassarre2019explainability} assessed feature importance by computing the gradient of the output with respect to each input feature; and generative XGNN \cite{yuan2020xgnn} seek to maximize prediction probability by generating an explanatory graph. Although these methods excel in node classification tasks, they often cannot control the model's focus on specific types of relationships during interpretation, which may lead to the model paying attention to less important structures, thereby providing low-quality explanations in CDR field.

\section{Methods}
\label{sec3}
In this section, we present the technical details of the proposed CETExplainer. We define CDR prediction as a link prediction task within a directed heterogeneous network, utilizing CETExplainer to provide node-level, fine-grained explanations for the prediction results, and ultimately constructing GT to evaluate CETExplainer.
\subsection{Directed graph structure construction}
\label{subsec3.1}
\begin{figure}[htbp]
    \centering
    \includegraphics[width=0.9\linewidth]{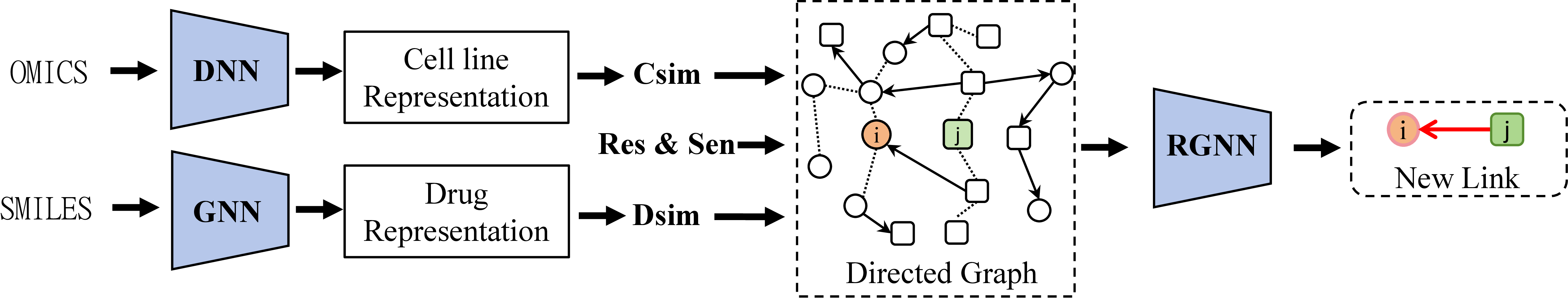}
    \caption{Process of Constructing a Directed Heterogeneous Network for Link Prediction. The multi-omics data of cell lines is input into a DNN layer to obtain cell line representations. Similarly, SMILES data is processed through a GNN layer to derive drug representations. Cell similarity triples and drug similarity triples are obtained using cosine similarity. These triples, together with resistance and sensitivity triples, form a directed heterogeneous network. This network is then input into an R-GCN layer to perform link prediction, enabling the identification of new links.}
    \label{fig:1}
\end{figure}

The process of constructing a directed heterogeneous network from raw data is shown in Figure \ref{fig:1}. Inspired by GraphCDR \cite{liuGraphCDRGraphNeural2022a}, we utilize DNN layers to integrate multi-omics data from cell lines to derive each cell line representations \( H_c \in \mathbb{R}^{cell \ num \times dim} \) and use GNN layers to extract drug representations \(H_d \in \mathbb{R}^{drug \ num \times dim}\) from drug SMILES.
For similarity triples, based on the node representation
\( H_c \) and \(H_d\), we apply cosine similarity with thresholds of 0.96 for cell lines and 0.85 for drugs to obtain cell line triples and drug similarity triples. These specific thresholds are chosen to ensure that similar cell line triples account for 20\% of all possible cell line triples and the same applies to drug triples.

For response data, we convert the half-maximal inhibitory concentration (IC50) values into sensitivity and resistance relationships, following the approach in \cite{pengPredictingDrugResponse2022}. A drug is considered sensitive in a cell line if its IC50 is below the sensitivity threshold; otherwise, it is considered resistant. By combining the similarity data and the response data, we construct a directed heterogeneous graph \(G=(\mathcal{V},\mathcal{E},\mathcal{R})\). Let \(\mathcal{V}\) denote the set of cell line nodes, \(\mathcal{E}\) denote the edges between these nodes, and \(\mathcal{R}\) represent the different types of relationships. In our project, there are four types of relationships: sensitivity (Sen), resistance (Res), drug similarity (Dsim), and cell similarity (Csim).

\subsection{Prediction method}
\label{subsec3.2}
To comprehend the underlying mechanisms of the CETExplainer algorithm, it is essential to have a foundational understanding of GNNs. Given that R-GCN has demonstrated its efficacy in handling downstream tasks based on multi-relational directed heterogeneous graphs \cite{schlichtkrull2018modeling}, we choose to utilize the R-GCN model to address the link prediction problem in the directed heterogeneous graph G and acquire new potential links, as shown in Figure \ref{fig:1}. 

In the R-GCN framework, parameter updates involve two main processes. First, for each node \(v_i\in \mathcal{V}\) at layer \(l\), the model aggregates features from itself and neighboring nodes \(v_j\in\mathcal{V}\) that are connected through relationship \(r\in \mathcal{R}\). This means that each relationship corresponds to a specific parameter \(W_r\). Second, the aggregated information from layer \(l\) undergoes a nonlinear transformation before being passed to the next layer \(l+1\), resulting in the final embedded features. The following is the propagation model for calculating the forward-pass update of a node denoted by \(v_i\): 
\begin{equation}
h_i^{\left(l+1\right)}=\sigma\left(\sum_{r\in\mathcal{R}}{\sum_{j\in\mathcal{N}_i^r}\frac{1}{c_{i,r}}W_r^{\left(l\right)}h_j^{\left(l\right)}}+W_0^{\left(l\right)}h_i^{\left(l\right)}\right)
\end{equation}
where \(r\) represents the type of relationship, \(N_i^r\) denotes the set of neighboring nodes of node \(i\) with relationship \(r\), \(c_{i,r}\) is a normalization constant specific to the problem, \(W_r\) is the parameter for relationship \(r\), \(h^{\left(l\right)}\in R^{d\left(l\right)}\) is the hidden state of the node at the \(l\)-th layer of the neural network, and \(d(l)\) is the dimensionality of the representation at that layer. 

We treat the triples that already exist as positive samples, and randomly perturb the subject node or object node then generated the negative samples. The technical details can be found in R-GCN \cite{schlichtkrull2018modeling}.
\subsection{Edge-type-specific interpretability algorithm}
\label{subsec3.3}
\begin{figure}[htbp]
    \centering
    \includegraphics[width=0.9\linewidth]{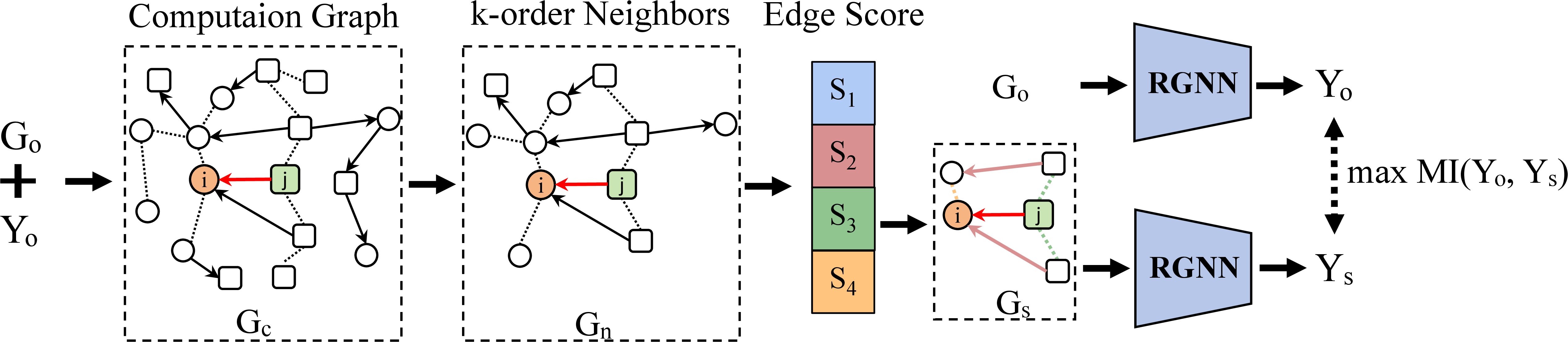}
    \caption{Main Architecture of CETExplaine. \(G_o\) represents the original directed heterogeneous graph, and the "new link" refers to the prediction result the link prediction task. \(G_c\) is the combination of Go and "new link". \(G_n\) is the subgraph form by \(k\)-hop neighborhood nodes and edges that extracted from \(G_c\). \(G_s\) is the explanation subgraph that derive from \(G_n\). \(Y_o\) and \(Y_s\) are the prediction results obtained by input \(G_o\) and \(G_s\) to RGCN layers. The objective is to guide the explanation model to prioritize crucial edges within the interpretive process.}
    \label{fig:2}
\end{figure}

In this section, we provide an intuitive description of the underlying principles of the CETExplainer algorithm as shown in Figure \ref{fig:2}. 
The original graph \(G_o\) and corresponding prediction result \(Y_o\) are combined to generate the computational graph \(G_c\). It should be noted that \(G_o\) is the directed graph and \(Y_o\) is new link in Figure \ref{fig:1}. Then, we extract the \(k\)-hop neighborhood graph \(G_n\) of the edge to be explained. We apply a optimizable mask and the edge-type-specific weighting mechanism to \(G_n\) to obtain the initial explanation subgraph \(G_s\). Subsequently, the mask is iteratively optimized by maximizing the mutual information between the \(G_s\) and the \(G_o\), ultimately deriving the final explanation subgraph \(G_s\).

CETExplainer optimizes both the mutual information and the structure scores simultaneously, facilitating the identification of significant subgraphs. it controls the model's attention on different relationships to select key explanatory subgraphs, thereby improving the model's explanatory capacity. Next, we formalize CETExplainer from two aspects: mutual information and structural scoring.

\subsubsection{Mutual information}
Given the prediction link \((i, r, j)\), CETExplainer identifies the associated cluster \(V_i\) for cell line \(i\) and \(V_j\) for drug \(j\), where \(V_i\) includes all cell line nodes that have similar or indirectly similar relationships with cell line \(i\), and \(V_j\) comprises all drug nodes similarly associated with drug \(j\). Within the graph \(G_c\) formed by these cell line and drug clusters, given a pair of nodes \(v_i\) and \(v_j\), the goal is to determine a subgraph \(G_s\) that is most critical for the prediction outcome \((i,r,j)\). Inspired by GNNExplainer \cite{ying2019gnnexplainer}, we use Mutual Information (MI) to formalize the concept of importance, and we formulate \(MI\) as follows:
\begin{equation}
    \underset{G_s}{max}\ MI\left(Y,G_s\right)=H\left(Y\right)-H\left(Y\middle| G=G_s\right)
\end{equation}
for node \(v\), \(MI\) quantifies the change in the probability of prediction when \(v\)’s computation graph is limited to explanation subgraph \(G_s\). Examining Eq. (2), we see that the entropy term \(H(Y)\) is constant because model parameter \(\Phi\) fixed for a trained GNN. Besides, conditional entropy \(H(Y|G=\ G_s)\) can be expressed as follows:
\begin{equation}
    H\left(Y\mid G=G_s\right)=-E_{Y\mid G_s}\left[\log{P_\Phi}\left(Y\mid G=G_s\right)\right]
\end{equation}
\subsubsection{Structure score}
To account for the varying contributions of different edges during the explanation process, we have augmented the Mutual Information (\(MI\)) with an additional optimization objective. This ensures that while maximizing \(MI\), the score of \(G_s\) is also maximized, thereby refining the model to favor specific edges when selecting \(G_s\) during the explanation. The scoring function is defined as follows:
\begin{equation}
    \text{Score}(G_s) = \sum_{r \in G_s}^{R}\sum_{i=0}^{N} w_r r_i + \text{Penalty}(N)
\end{equation}
it is computed based on the edges within the subgraph \(G_s\), where \(N\) is the total number of edges in \(G_s\), \(r\) denotes a specific type of relationship within the subgraph, \(N_r\) represents the number of nodes involved in relationship \(r\),\(w_r\) is the predefined weight for edges of type \(r\), and \(r_i\) denotes a node involved in the relationship \(r\). A ``Penalty'' coefficient, related to the number of edges, is used to control the size of the graph. The penalty is defined as: 
\begin{equation}
    \text{Penalty}(N)=\alpha N+\beta N^2
\end{equation}
where \(\alpha\) and \(\beta\) are positive hyperparameters used to adjust the weight of the penalty term.By synthesizing this score with equations (1) and (2), the final optimization objective is as follows: 
\begin{equation}
    \underset{G_s}{max} MI(Y,(G_s,X_s))+Score(G_s)
\end{equation}
\subsection{Evaluation algorithm for explanations}
\label{subsec3.4}
In the previous section, we introduced our interpretable model, CETExplainer. A natural question is how to evaluate the explanations it provides. In this part, we propose a quantitative evaluation method based on GT on the heterogeneous network constructed from both similarity and drug response data. We will elaborate on this method from two aspects: constructing the GT and assessing the metrics.
\subsubsection{Constructing ground truth}
\begin{figure}[htbp]
    \centering
    \includegraphics[width=0.8\linewidth]{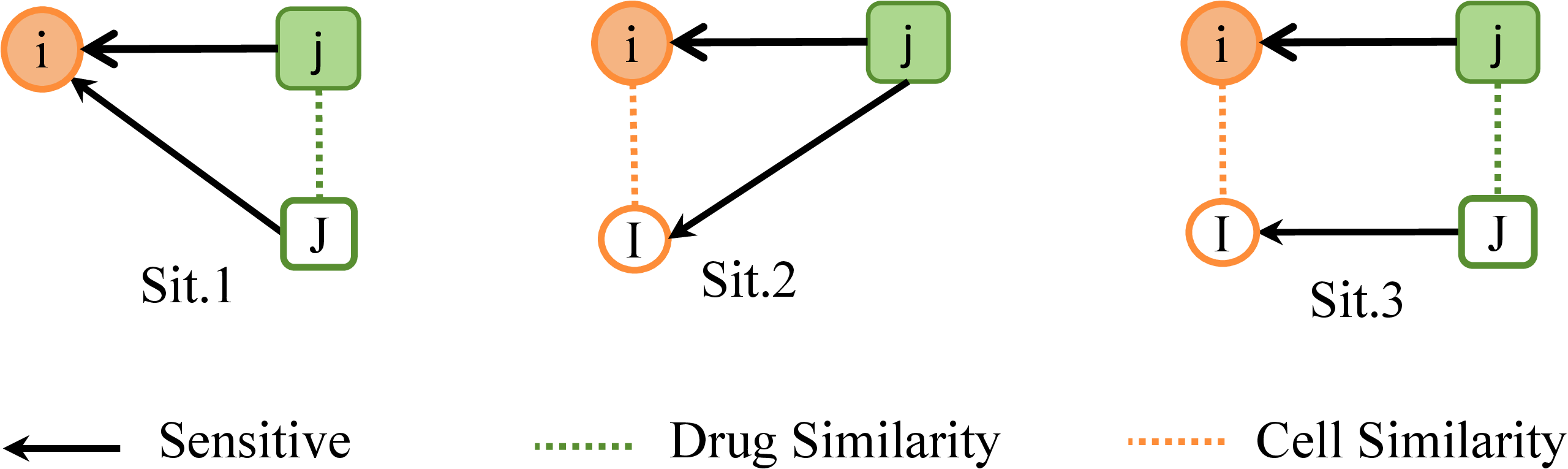}
    \caption{Three situations for constructing GT. Sit. 1 demonstrates that if \(i\) is sensitive to \(J\), and \(J\) is similar to \(j\), then \(i\) is sensitive to \(j\). Sit. 2 shows that if \(I\) is sensitive to \(j\), and \(I\) is similar to \(i\), then \(i\) is sensitive to \(j\). Sit. 3 demonstrates that if \(I\) is sensitive to \(J\), \(I\) is similar to \(i\), and \(J\) is similar to \(j\), then \(i\) is sensitive to \(j\).}
    \label{fig:3}
\end{figure}

The directed heterogeneous network comprises resistance edges, sensitivity edges, drug similarity edges, and cell line similarity edges. The final explanation can be derived from these edges and corresponding nodes. We utilize these relationships and known triples to construct the GT. Initially, we identify the one-hop similar cell line neighbors and one-hop similar drug neighbors. Subsequently, we search for existing sensitive or resistant relationships among these similar drugs and cell lines to formulate the GT. The final GT we constructed can be categorized into three scenarios, as illustrated in Figure \ref{fig:3}.Given an edge to be explained, by comparing the overlap between the triples provided by the interpretable model and the GT, we can provide a quantitative assessment for explanations.

\subsubsection{Evaluating metrics}
To quantitatively assess the quality of the model explanations using GT, we have introduced three evaluation metrics: Precision@k, Recall@k and F1@k. Precision@k is used to measure the proportion of correct explanations within the top-k ranked predictions, while Recall@k assesses the proportion of these correct top-k explanations relative to the total number of correct explanations provided by the model. The computational methods for these metrics are defined as follows: 
\begin{equation}
    \mathrm{Precision@k}=\frac{\mathrm{TP@k}}{\mathrm{TP@k}+\mathrm{FP@k}}
\end{equation}

\begin{equation}
    \mathrm{Recall@k}=\frac{TP@k}{\mathrm{TP@k}+\mathrm{FN@N}}
\end{equation}

\begin{equation}
    F1@k=2\cdot\frac{\mathrm{Precision@k}\cdot\mathrm{Recall@k}}{\mathrm{Precision@k}+\mathrm{Recall@k}}
\end{equation}
where \(\mathrm{TP@k}\) represents the number of correct explanations within the top-k predictions made by the model, \(\mathrm{FP@k}\) refers to the number of incorrect explanations within these top-k predictions, and \(\mathrm{FN@N}\) indicates the true instances that were not correctly explained among the N explanations provided by the model. In addition, to evaluate how the model is influenced by the convergence of the prediction model, we define a metric called explanation stability. This metric measures stability by calculating the overlap of explanations provided by the explanation model at different epochs of the prediction model as defined below:
\begin{equation}
\mathrm{Stability}\left(\mathrm{T}_{\mathrm{e}_\mathrm{1}},\mathrm{T}_{\mathrm{e}_\mathrm{2}}\right)=\frac{\sum_{i=1}^{N_E}{{NUM(T}_{e_1}^i\cap T_{e_2}^i)}}{N_E}
\end{equation}
where \(T_{e_k}^i\) denotes the explanation provided by the explanation model for the edge \(i\) to be explained when the prediction model is at epoch \(e_k\). \(N_E\) represents the total number of edges to be explained.
These metrics collectively assess the model’s accuracy and its coverage of actual scenarios, serving as critical standards for evaluating the explainability of the model. 
\section{Experiments and results}
\label{sec4}
In this section, we first introduce the experimental setups \footnote{Code:\href{https://github.com/AhauBioinformatics/CETExplainer}{https://github.com/AhauBioinformatics/CETExplainer}.} and then demonstrate the performance of the proposed model CETExplainer through comparison with baseline methods. 

\subsection{Datasets}
\label{subsecc4.1}
Drug response triples are derived from the Genomics of Drug Sensitivity in Cancer (GDSCv2) \cite{yang2012genomics}, a comprehensive resource that compiles sensitivity response data of tumor cell lines to various drugs, which is the largest public repository for tumor cell drug sensitivity currently available. The features of cell lines are primarily sourced from the Omics data of the Cancer Cell Line Encyclopedia (CCLE) \cite{barretina2012cancer}, while the drug features are derived from SMILES data hosted on PubChem \cite{kim2019pubchem}. In total, we utilize 17,071 drug response pairs, covering 477 cell lines and 157 drugs. After feature processing, we obtained 49,343 cell similarity triples and 5,540 drug similarity triples. Ultimately, our data include four types of triples: resistance, sensitivity, drug similarity, and cell similarity. 

Based on these data, we constructed the GT. Specifically, of the edges to be explained, 33.01\% have a GT count of fewer than 1,009, which means that most of the edges to be explained have a corresponding GT count between 10 and 1,009. The specific distribution is shown in Table \ref{tab:1}.
\begin{table}[htbp]
\centering
\caption{Distribution of GT Numbers}
\label{tab:1}

\begin{tabular}{c c c}
\hline
\textbf{GT Num} & \textbf{Edge Count} & \textbf{Proportion (\%)} \\ 
\hline
10-1009        & 1088                & 33.01                    \\ 
1010-2009      & 196                 & 5.95                     \\ 
2010-3009      & 286                 & 8.68                     \\ 
3010-4009      & 177                 & 5.37                     \\ 
4010-5009      & 260                 & 7.89                     \\ 
5010-6009      & 227                 & 6.89                     \\ 
6010-7009      & 341                 & 10.35                    \\ 
7010-8009      & 296                 & 8.98                     \\ 
8010-9009      & 263                 & 7.98                     \\ 
9010-10009     & 146                 & 4.43                     \\ 
10010-11009    & 16                  & 0.49                     \\ 
\hline
\end{tabular}

\end{table}
\subsection{Implementation details}
\label{subsecc4.2}
In this study, five-fold cross-validation is employed to assess the performance of CETExplainer. Specifically, we combine the drug response triples with the similarity triples, shuffle them randomly, and then evenly divide them into five subsets. Each subset is used in turn as a test set, with the remaining subsets serving as the training set. The performance metrics are averaged to produce the final results. The ratio of positive to negative samples in the drug response prediction model is set at 1:2.

Regarding the weight scores of edges in CETExplainer, we referenced the distribution of unweighted edge types. Noticing that Relationships 1 and 2 received less attention, we set the weight scores for the four types of edges to 0.1, 0.4, 0.4, and 0.1, respectively, to balance the model’s focus on various edge types as much as possible.

\subsection{Experimental results}
\label{subsecc4.3}
Given the scarcity of directly applicable explanation algorithms for link prediction tasks, we selected the perturbation-based GNNExplainer and the counterfactual-based ExplaiNE as baseline models. We began by comparing the explanation metrics of the baseline models, followed by an analysis of these metrics in relation to the proportion of different edge types within the explanations.
Subsequently, we identified the optimal epoch for CETExplainer through parameter tuning. Additionally, we performed further experiments to investigate how the number of training epochs in the prediction model influences the specificity of the model's explanations.

\subsubsection{Comparison with baselines}
\begin{figure}[htbp]
    \centering
    \subcaptionbox{Comparison of Metrics for Baseline Models}{\includegraphics[width=0.48\textwidth]{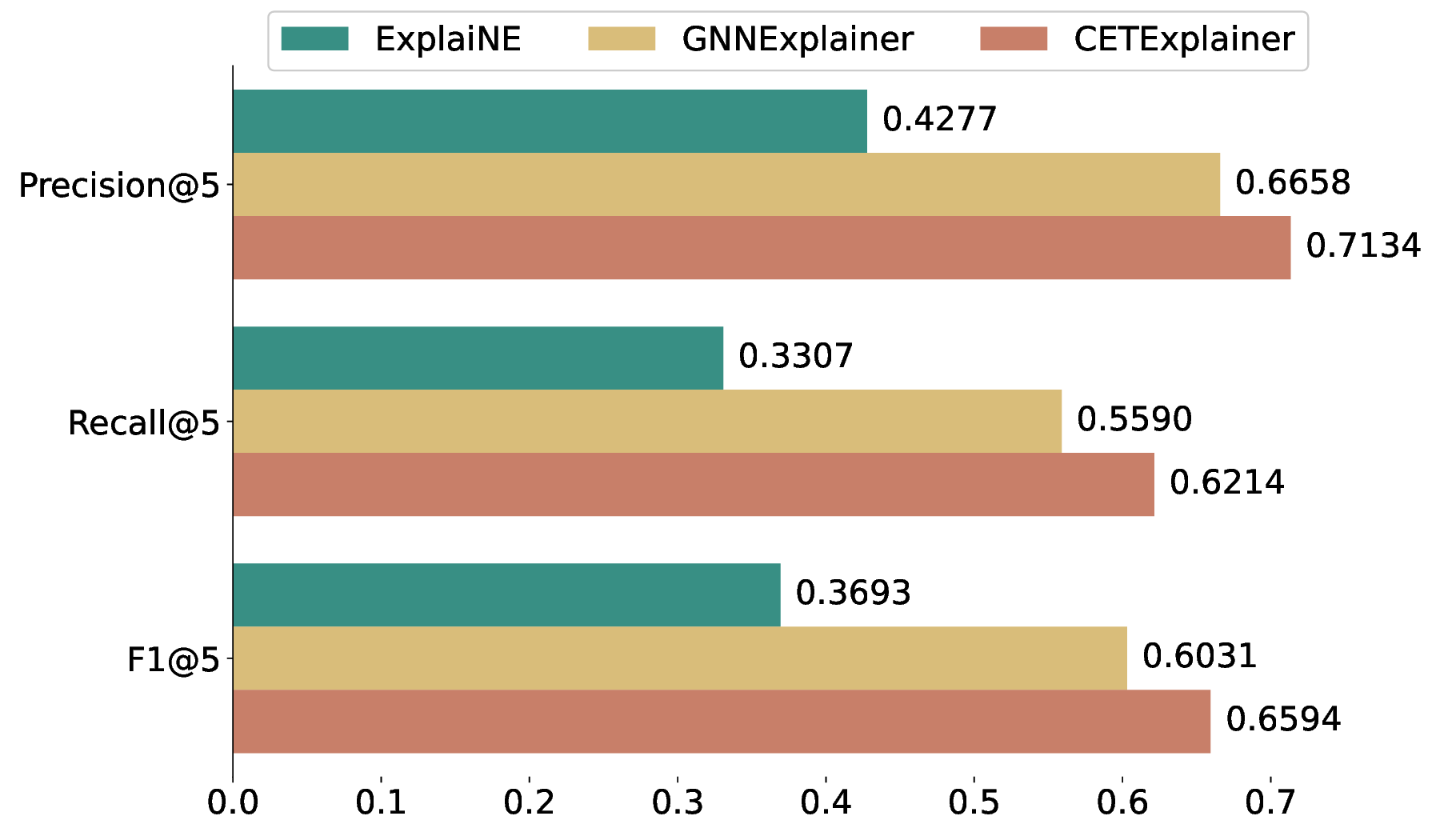}}
    \hfill
    \subcaptionbox{Distribution of Edge Types in Explanations}{\includegraphics[width=0.48\textwidth]{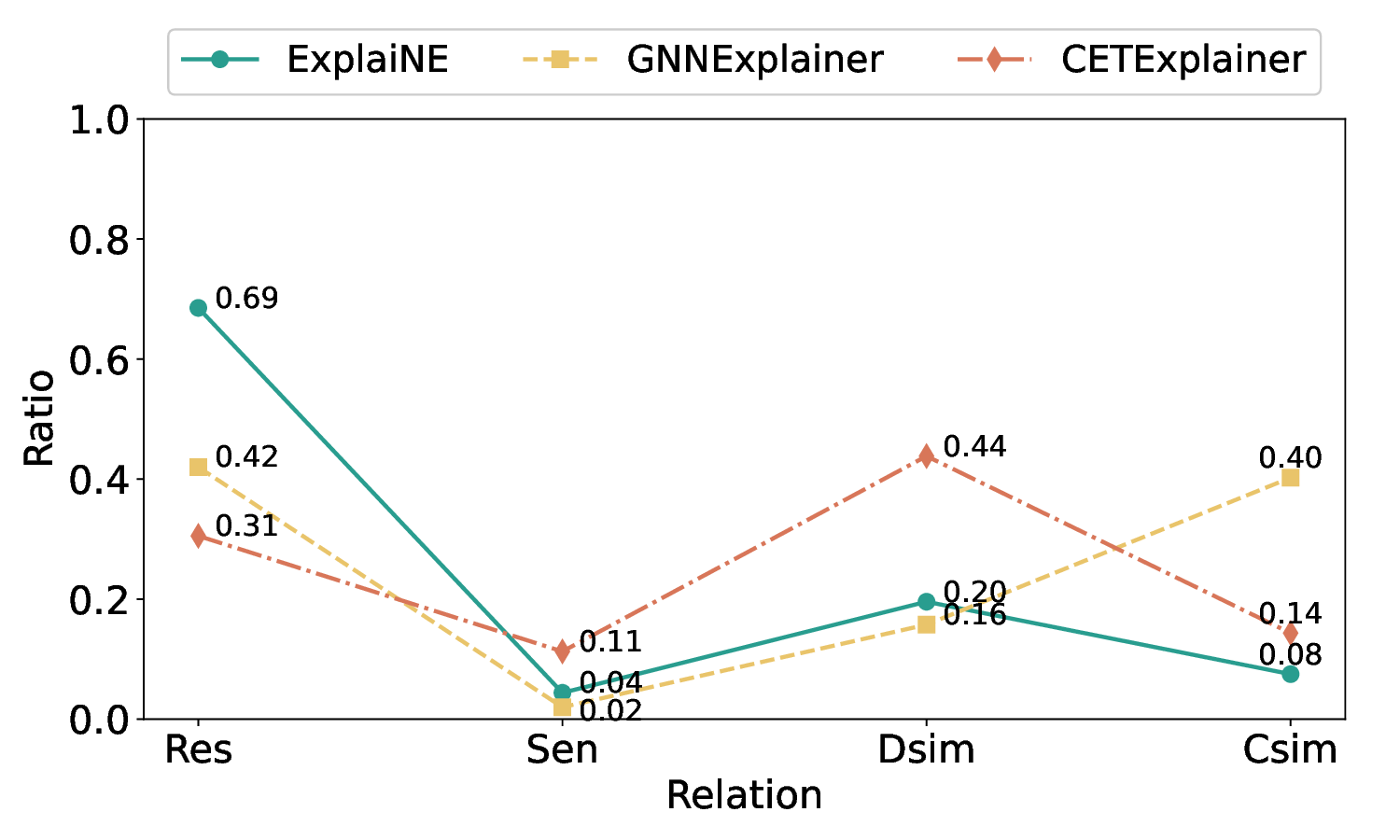}}
    \caption{(a) Results showing the quantitatively evaluate CETExplainer using three proposed metrics and compare it with the baseline models, ExplaiNE and GNNExplainer.(b) Res represents resistance, Sen represents sensitivity, Dsim represents drug similarity, and Csim represents cell similarity. CETExplainer improves the proportion of sensitivity, which is more important for explanations and predictions compared to the other two models, and maintains a more balanced distribution of various edge types. This results in an improved quality of the explanations.}
    \label{fig:quantify eval}
\end{figure}

Figure \ref{fig:quantify eval}(a) presents a comparison between CETExplainer and two baseline methods, GNNExplainer and ExplaiNE. The metrics for all models were obtained under conditions with optimal parameter settings. It is evident that CETExplainer achieves the best performance with an F1 score of 0.6594. Notably, although GNNExplainer also employs mutual information maximization to optimize the mask, it faces challenges due to the imbalance in edge types within the dataset. If it fails to specifically focus on certain edge types, it can lead to insufficient explanations for some edges due to the sparsity issues. As shown in Figure \ref{fig:quantify eval}(b), we calculated the proportion of various edge types within the top 10 explanations for each edge. In GNNExplainer's explanations, the proportion of type 1 edges is only 0.02, whereas CETExplainer increases the weight of type 1 edges, raising their proportion to 0.11 and thereby significantly improving accuracy. The subpar performance of ExplaiNE in this context may be attributed to its explanation method, which primarily assesses the impact of removing a single edge on the prediction. Since cancer drug response is highly complex, the removal of a single edge does not significantly affect the overall prediction. Therefore, ExplaiNE is not suitable for data with high inter-node correlations.
\begin{figure}[htbp]
    \centering
    \includegraphics[width=\linewidth]{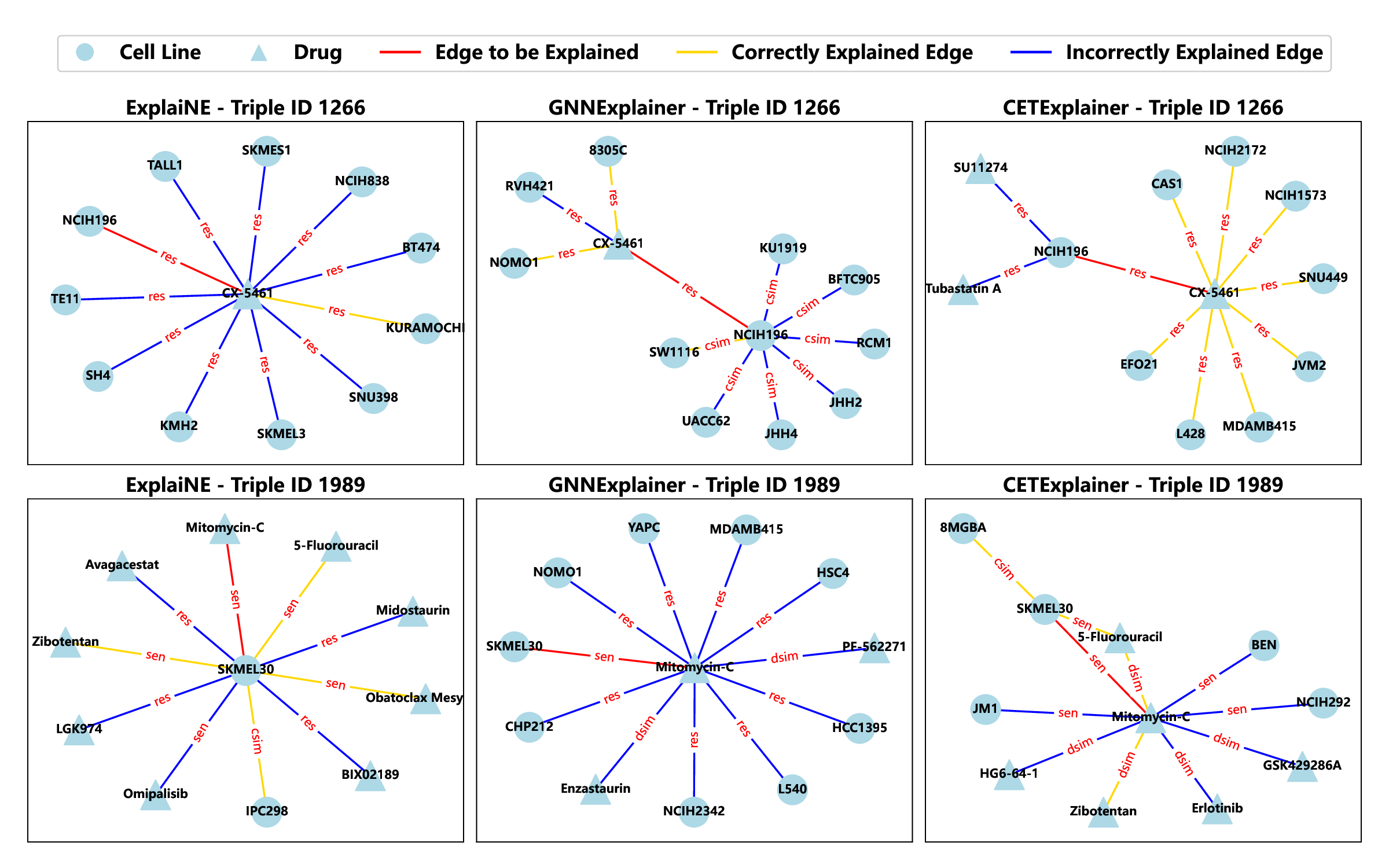}
    \caption{Qualitative evaluation of the three explanation models on triplet instances 1266 and 1989.}
    \label{fig:casestudy}
\end{figure}

\subsubsection{Qualitative Assessment}
We perform the qualitative assessment on our proposed method and the baseline models. We visualize the fine-grained explanation of the predicted link as is shown in Figure \ref{fig:casestudy}. 

For the res edge to be explained (CAL29, res, NSC-207895), both CETExplainer and GNNExplainer correctly predict 9 edges, which is more than ExplaiNE's 4 edges. Our model successfully identifies the complete GT structure shown in Figure 3. For the sen edge to be explained (Mitomycin-C, sen, SKMEL30), our model and ExplaiNE both correctly predict 4 edges, while GNNExplainer fails to predict any edges correctly. Additionally, our model also identifies the GT structure.

Furthermore, the subgraphs generated by ExplaiNE and GNNExplainer for explanation purposes contain many res edges, which are difficult to directly interpret for sen edges based on node similarity. While the algorithms consider these resistance edges as contributing significantly to the model’s predictions, they lack biological interpretability. This fails to achieve the goal of explainability—generating a human-understandable explanation for the model’s prediction results.

\subsubsection{Explanation across epochs}
\begin{figure}[htbp]
    \centering
    \subcaptionbox{Explanation performance}{\includegraphics[width=0.48\textwidth]{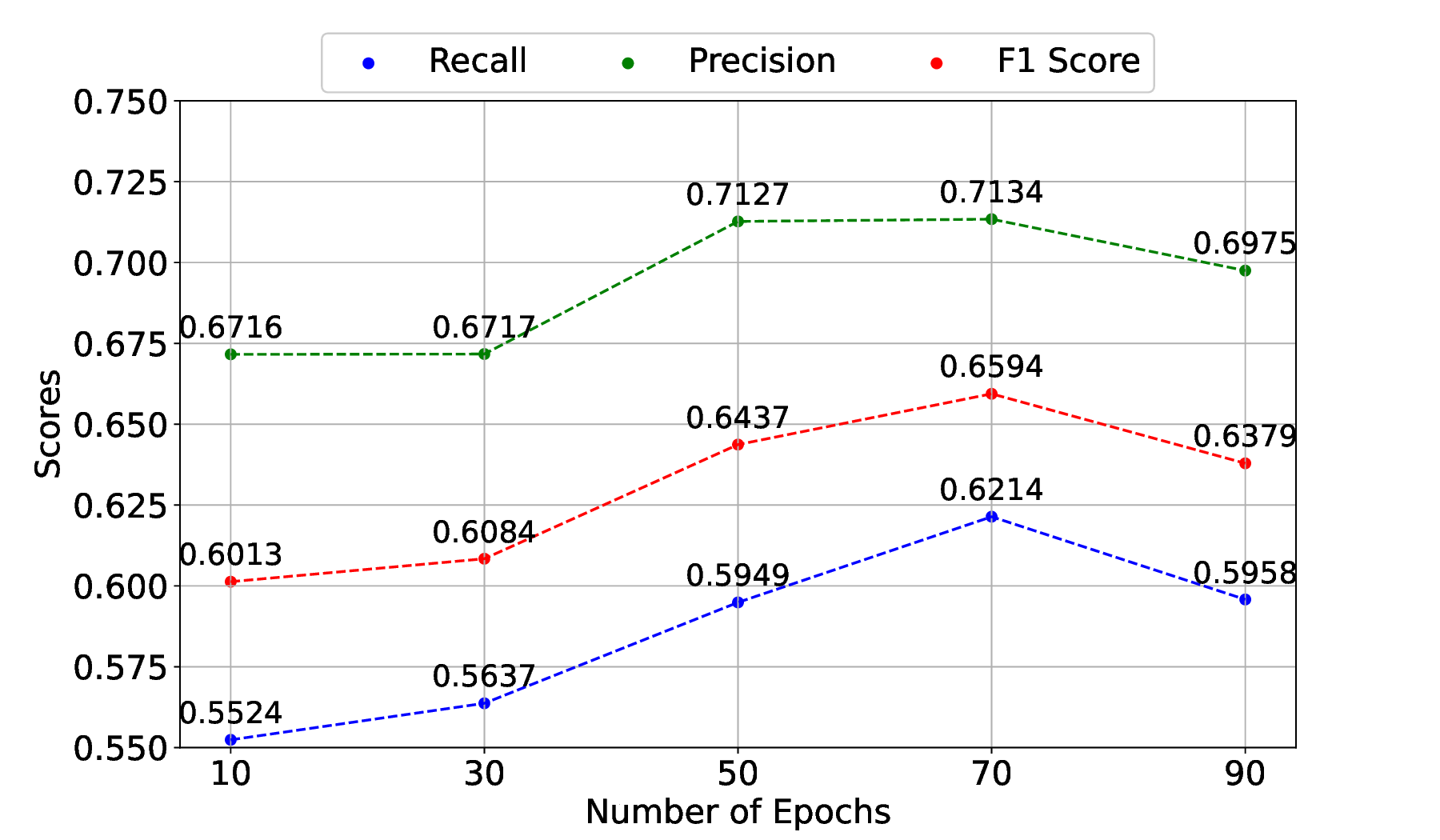}}
    \hfill
    \subcaptionbox{The overlap in explanations}{\includegraphics[width=0.48\textwidth]{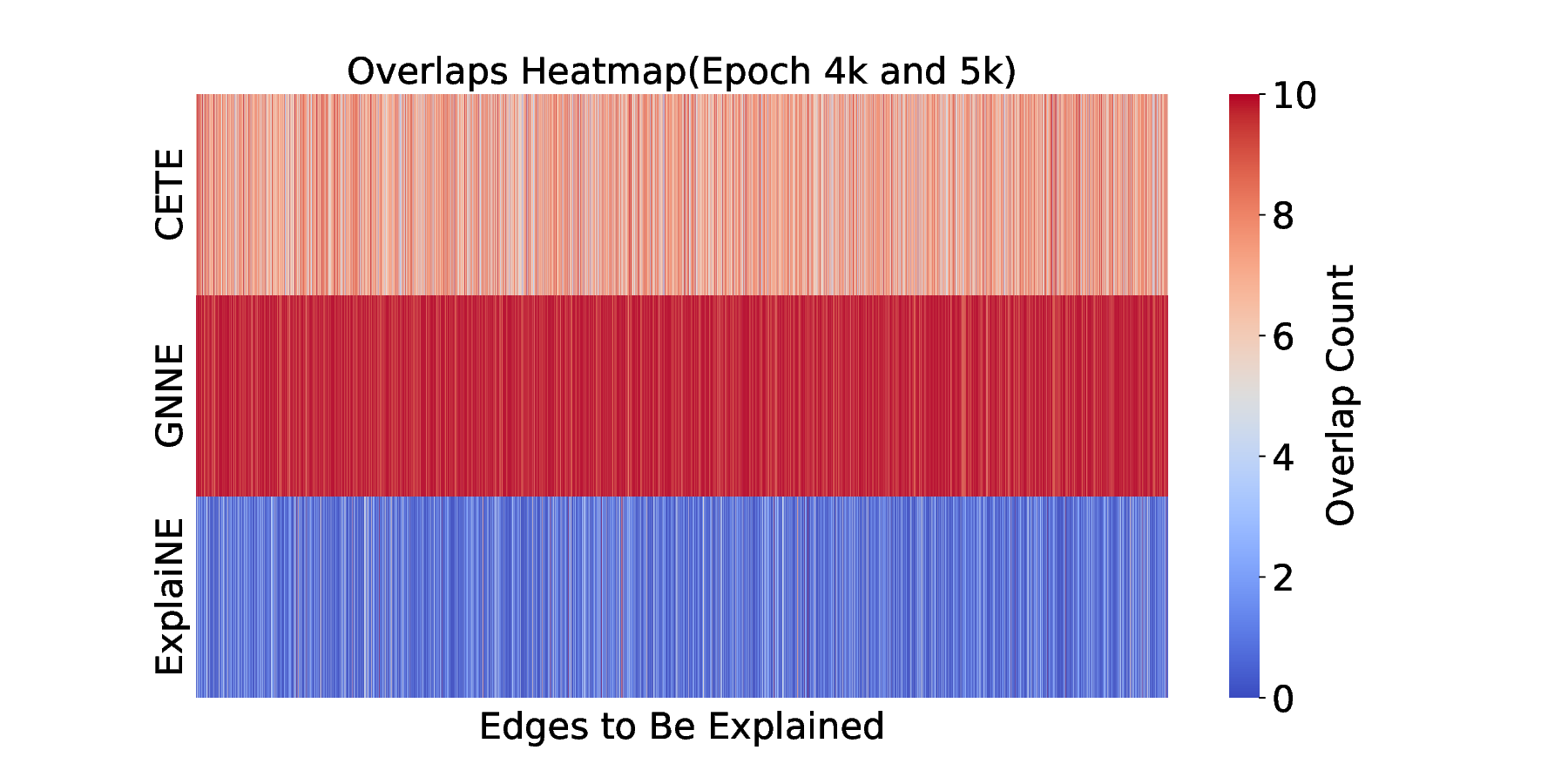}}
    \caption{(a)Results showing the explanation metrics of CETExplainer across different epochs. (b)Results showing the heatmap of the overlap in explanations provided by different interpretability models for prediction models trained at epochs 4000 and 5000, where "GNNE" represents GNNExplainer and "CETE" represents CETExplainer.}
    \label{fig:5}
\end{figure}

To evaluate the stability of the explanation model, we analyzed how the explanation metrics changed across different training epochs. We trained models with the same architectures as previously described, varying the number of training epochs from 10 to 90. Each experiment utilized two layers of RGCN. The results, as shown in Figure \ref{fig:5}(a), indicate that the explanation model achieves optimal performance at 70 epochs, with an F1 score of 0.6594.
\subsubsection{Stability of the explanation model}
After analyzing the metrics of the explanation model across different epochs, we aim to investigate how the number of training epochs of the prediction model influences the explanations provided by the explanation model. We refer to the influence as model stability and compared the stability of three different models. Using data from the second fold as an example, we calculated the explanation overlap under adequate training conditions (4k and 5k epochs). This overlap is defined as the average number of overlapping explanations per edge. Additionally, we presented a heatmap of the overlap, as shown in Figure \ref{fig:5}(b). 
The results demonstrate that the average stability of our model and GNNExplainer both exceed 8.0, while ExplaiNE exhibits a stability of only 1.058. This indicates that ExplaiNE is more susceptible to variations in the prediction model's training conditions.

\section{Conclusion}
\label{sec5}
In this paper, we propose CETExplainer, a novel interpretability algorithm with controllable edge weights for multi-relational GNNs, tailored for explaining drug response predictions. CETExplainer leverages a learning framework that maximizes mutual information and the structure scores of explanatory subgraphs, thereby achieving fine-grained, high-quality explanations. Specifically, we integrate multi-omics features of cell lines and encode drug SMILES to learn representations of cell line and drug nodes effectively. Additionally, we create GT based on existing drug response data, enabling non-experts to comprehend and quantitatively assess the effectiveness of our explanation model. Extensive experimental results demonstrate that CETExplainer surpasses two baseline models in providing interpretable explanations for drug response predictions. In the future, we need to further investigate the relationship between prediction and explanation, as well as how explanations can enhance prediction performance.
\section{Funding}
This work was supported by the grants from the National Natural Science Foundation of China (62102004), the Introduction and Stabilization of Talent Project of Anhui Agricultural University (yj2019-32).

\bibliographystyle{elsarticle-num-names.bst}
\bibliography{CETE.bib}

\end{document}